\documentclass[letterpaper, 10 pt, conference]{ieeeconf}  

\IEEEoverridecommandlockouts                              

\usepackage{epsfig} 
\usepackage{amsmath} 
\usepackage{amssymb}  
\usepackage{algorithm}
\usepackage{nccmath}
\usepackage[noend]{algpseudocode}
\usepackage{graphicx}
\usepackage{caption}
\usepackage{adjustbox}
\usepackage{grffile}
\usepackage{grffile}
\usepackage{subcaption}
\usepackage{float}
\usepackage{mathtools}

\title{\LARGE \bf
Towards Physically Safe Reinforcement Learning under Supervision}

\author{Yinan Zhang$^{1}$, Devin Balkcom$^{2}$ and Haoxiang Li$^{3}$ 
\thanks{$^{1}$Department of Computer Science, Dartmouth College, Hanover, NH 03755, USA.
        {\tt\small yinan.zhang.gr@dartmouth.edu}}%
\thanks{$^{2}$Department of Computer Science, Dartmouth College, Hanover, NH 03755, USA.
        {\tt\small devin.balkcom@dartmouth.edu}}%
\thanks{$^{3}$Adobe Research, Adobe System Inc., San Jose, CA 95110, USA.
        {\tt\small haoxli@adobe.com}}%
}

\begin{document}

\maketitle
\thispagestyle{empty}
\pagestyle{empty}

\begin{abstract}
This paper addresses the question of how a previously available control policy $\pi_s$  can be used as a supervisor to more quickly and safely train a new learned control policy $\pi_L$ for a robot. A weighted average of the supervisor and learned policies is used during trials, with a heavier weight initially on the supervisor, in order to allow safe and useful physical trials while the learned policy is still ineffective. During the process, the weight is adjusted to favor the learned policy. As weights are adjusted, the learned network must compensate so as to give safe and reasonable outputs under the different weights. A pioneer network is introduced that pre-learns a policy that performs similarly to the current learned policy under the planned next step for new weights; this pioneer network then replaces the currently learned network in the next set of trials. Experiments in OpenAI Gym demonstrate the effectiveness of the proposed method.

\end{abstract}


\section{Introduction}

In the early stages of training, outputs from a learning policy can be unreasonable and lead to catastrophic failures. Motivated by the twin goals of better data efficiency and fewer failures in learning-based control, we propose an algorithm to leverage a possibly imperfect external supervisor policy to help accelerate the learning, and introduce the concept of a pioneer policy to realize safe progressive updates of the learner policy. In our framework, the training signals come from both the interactions with the environment and the supervisor policy, and the goal is that eventually the learner policy surpasses or at least matches the performance of the supervisor policy.


Consider a problem of robot control in which a traditional manually designed control policy is available and effective. However, new un-modeled conditions may arise: a drone may encounter shifting wind patterns or pick up an unsteady load for manipulation, or a self-driving vehicle may encounter unfamiliar lighting conditions or slick roads. The goal of this paper is to use a previously available control policy $\pi_s$ to more safely and effectively learn a new control policy $\pi_L$ using deep reinforcement learning. Once a policy is learned, it may in turn be used as a supervisor to bootstrap the learning of better policies as training data becomes available.

Define $\pi_c$ to be the current policy at a particular time, and compute it as a weighted linear combination of $\pi_s$ and $\pi_L$:

\begin{equation}
\pi_c = k \pi_s + (1 - k) \pi_L.
\end{equation}

Initially, $k$ is set close to $1$, so the resulting control policy is dominated by the supervisor. Trials are conducted and $\pi_L$ is learned with the objective of improving $\pi_c$ with regards to the current task. Once a target score has been achieved, $k$ is reduced by some percent, for example, 4\%.

Notice that if $k$ is reduced suddenly, the policy $\pi_c$ also changes suddenly at any point where $\pi_s$ and $\pi_L$ indicate different controls. This may cause a robot to take surprising and perhaps dangerous actions. To avoid this issue, during training of $\pi_L$, we also train a {\em pioneer} policy $\pi_p$ simultaneously, but for a different weighting of the supervisor $k'$: $k' \pi_s + (1 - k') \pi_p$. In order resolve the discontinuity, the pioneer network is trained with the goal that:
\begin{equation}
\pi_c = k' \pi_s + (1 - k') \pi_p.
\end{equation}

Once $\pi_p$ and $\pi_L$ both reach desired objectives, then $k$ is updated and $\pi_L$ is replaced by the pioneer network $\pi_p$.


\begin{figure}[tp]
    \begin{subfigure}[b]{0.24\textwidth}
    \centering
    \includegraphics[scale=1.1]{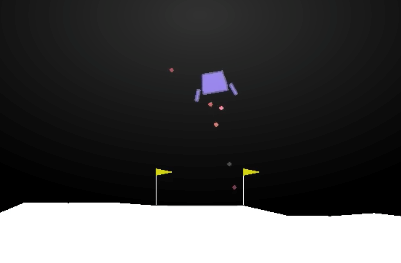}
    \caption{LunarLanderContinuous-v2 environment}
    \label{fig:lunar}
    \end{subfigure}
    \begin{subfigure}[b]{0.24\textwidth}
    \centering
    \includegraphics[width=1.3in]{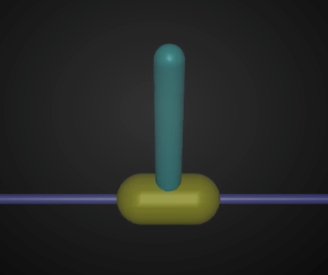}
    \caption{InvertedPendulum-v1 environment}
    \label{fig:pendulum}
    \end{subfigure}

			\begin{subfigure}{0.24\textwidth}
    \centering
    \includegraphics[width=1.2in]{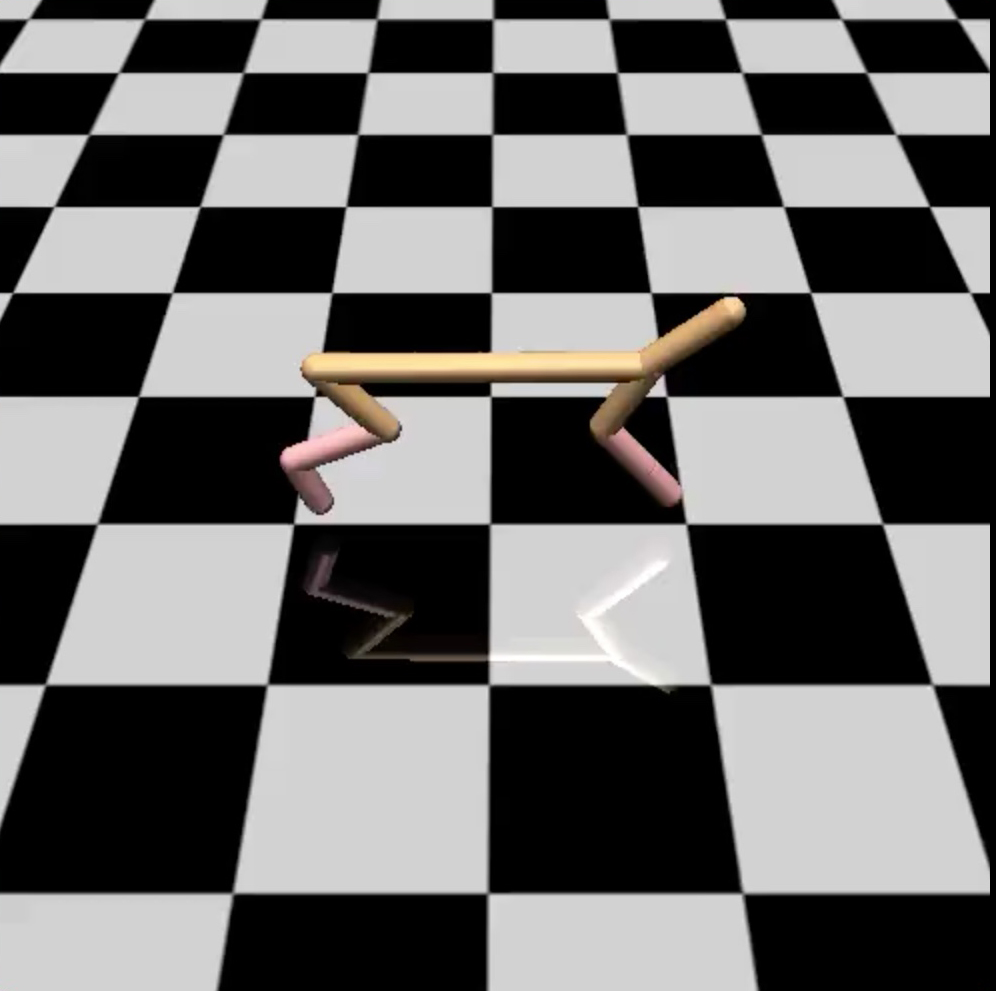}
    \caption{HalfCheetah-v1 environment}
    \label{fig:half_cheetah}
    \end{subfigure}
	 \begin{subfigure}{0.24\textwidth}
    \centering
    \includegraphics[scale=0.4]{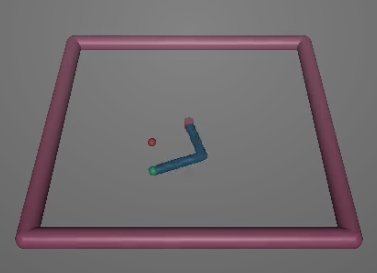}
    \caption{Reacher-v1 environment}
    \label{fig:reacher}
    \end{subfigure}

		\caption{OpenAI gym environments used as experiment tasks.}
		\label{fig:envs}
\end{figure}

\section{Related Work}


Recent progress in deep learning has led to several success stories in various domains~\cite{mnih2015human,NIPS2012_4824,levine2016end}.
Nevertheless, several practical issues for learning-based control remain under-explored, including efficiency and safety of the training process.




\textbf{Supervised learning.} Supervised learning is the machine learning task of inferring a function from labeled training data~\cite{mohri2012foundations}, finding a mapping between the input object and a desired output value while minimizing an error signal. Linear regression and multivariate linear regression~\cite{seber2012linear}, for instance, aim at finding a linear relationship between variables. Logistic regression~\cite{hosmer2013applied,kleinbaum2010analysis} considers specifically mapping the independent input variables to a binary output, while Naive Bayes classifiers~\cite{murphy2006naive,rish2001empirical} map the input to multiple categories.
Support Vector Machines (SVM)~\cite{suykens1999least,cortes1995support} are efficient large-margin classifiers for both linear and non-linear classifications.
Recently, deep neural networks have been widely and successfully adopted for many supervised learning tasks.

Supervised learning of an effective control policy requires a large number of state-action pairs, which are impractical to manually annotate. Imitation learning, which is generally supervised learning, works with an expert policy to query
optimal action given a state. To sample state-action pairs to train a policy with supervised learning, one widely used method is DAGGER~\cite{ross2011reduction}. DAGGER iteratively collects trajectories using the current policy
and then queries the expert policy at states along the trajectories for optimal actions to augment the training data to update the policy. DAGGER cannot produce a policy better than the expert policy. Further, in practice, it is not always possible to obtain an expert policy. In our work, we assume access to a supervisor policy, which may not be optimal, but provides reasonable performance. In the current work, the trained policy may provide better performance than the supervisor.


\textbf{Reinforcement learning and Deep reinforcement learning.}
 Reinforcement learning (RL) is often applied to problems involving decision making and maximizing feedback rewards as a performance signal. Markov decision processes (MDP)~\cite{puterman2014markov} provide a framework for modeling decision making. Watkins~\cite{watkins1989learning} proposed the Q-learning algorithm as a model-free technique for solving finite MDP problems. Q-learning works by learning a state-action value function that predicts the expected reward given an action at a state.

Deep neural networks have been introduced to reinforcement learning (DRL) for training policies from end-to-end to solve complex problems. A recent break through is the DQN from Minh {\em et al.}~\cite{mnih2015human,mnih2013playing}, in which neural networks are used to approximate the Q function for high dimensional state space. Levine {\em et al.}~\cite{levine2016end} developed an end-to-end DRL method that maps raw image input to robot motor controls. Silver {\em et al.}~\cite{silver2016mastering} combined tree search and DQN to beat the best human Go player. For continuous control, Lillicrap~\cite{lillicrap2015continuous} combined deep Q-learning with an actor-critic framework to learn a deterministic policy (DDPG).

Most of the DRL methods require a huge amount of training data to find a good policy. As the community is aware of this issue, many methods have been proposed to accelerate training. Popular techniques include adopting the advantage function from Schulman {\em et al.}, adding auxiliary tasks~\cite{jaderberg2016reinforcement} and asynchronous methods~\cite{mnih2016asynchronous}.
These methods are complementary to our proposed solution, with which we can potentially further accelerate the RL training.

We approach this problem by introducing a supervisor policy in training. We assume the existence of a reasonably good policy as a supervisor during training.
The learner policy learns from both the RL training signals and the supervisor policy. We observe significant acceleration of the learning process with respect to learning without a supervisor.
A relevant work from Hester {\em et al.}~\cite{hester2017learning} also propose to add supervised loss in RL training. However, their method does not address the safety issue
in RL training. Although the leaner policy improves quickly in generally in that work, there are sometimes unexpected behaviors, which may lead to catastrophic failures of the overall system.

\textbf{Safe reinforcement learning.}
{\em Safe reinforcement learning} highlights the importance of ensuring a reasonable system performance while searching for a new and improved policy~\cite{thomas2015safe}. We refer readers to~\cite{garcia2015comprehensive} for a more comprehensive literature survey.

Quite recently, Kahn {\em et al.}~\cite{kahn2017uncertainty}, proposed an uncertainty measure used to control the speed of the robot during learning to avoid collision.
For some tasks, e.g., stabilizing a quadrotor, a smaller magnitude of actions may not make the training safe. Our framework is general, and not task-specific.

\section{Background}
\label{sec:background}
In this section, we review the mathematical background of Q-learning and the technique of applying an artificial neural network to represent the actor-value function, including its extension to a continuous action space. Techniques discussed in this section will be applied in the next section where we combine a supervisor with a learning actor.

In a standard reinforcement learning setup, an agent interacts with an environment $E$ in discrete time-steps. Let $s_t$ be the observation of the environment at time $t$, $a_t$ be the action taken, and $r_t$ be a scalar reward feedback. We consider the action to be real valued $a_t \in \mathbb{R}^N$ and the environment is fully observed.

A {\em policy} $\pi:S\to \mathcal{P}(A)$ maps the state space $S$ to a probability distribution over the action space $A$. A policy determines the behavior of an agent. In the continuous action space, an {\em actor} function $\mu: S\to A$ is a {\em policy} that deterministically maps a state to a specific action. The reward of an action taken at a state is described by $r(s_t, a_t)$. The sum of discounted future rewards $R_t = \sum^T_{i=t}\gamma^(i-t)r(s_t, a_t)$ is the return from a state, where $\gamma \in [0,1]$ is a constant discount factor. 

The action-value function, or {\em Q function}, describes the expected return in state $s_t$ after taking an action $a_t$ and thereafter under policy $\pi$:
\begin{equation}
	Q^\pi = \mathbb{E}[R_t | s_t, a_t].
\end{equation}

Q-learning~\cite{watkins1989learning} uses a greedy policy to determine the action under current state that maximizes the return (the Q value).

The Bellman equation, as a necessary condition for optimality, is widely used to represent Q function in a recursive manner:

\begin{equation}
\label{eq:bellman}
	Q^\pi = \mathbb{E}_{r_t, s_{t+1}}[r(s_t, a_t) + \gamma \mathbb{E}_{a_{t+1}}[Q^\pi(s_{t+1}, a_{t+1})]].
\end{equation}

If both $S$ and $A$ are discrete, the Q function, a map of $S$ to $A$ can be described using a table. When $S$ is continuous, Deep Q networks (DQN) construct a network to represent the Q function. Let $\theta^Q$ be the parameter of $Q^\pi$ network. We optimize $\theta^Q$ by minimizing the loss:

\begin{equation}
\label{eq:Q_loss}
	L(\theta^Q) = \mathbb{E}[ (Q(s_t, a_t|\theta^Q) - y_t )^2 ],
\end{equation}
where
\begin{equation}
	y_t = r(s_t, a_t) + \gamma Q^\pi(s_{t+1}, \pi(s_{t+1}))
\end{equation}
is the observed return. Practically, the dependence on $y_t$ on parameters $\theta^Q$ is frequently ignored; we do the same. By iteratively updating the parameters, DQN predicts the return more and more accurately, thus better actions will be chosen.

DQN works only for discrete action spaces. In the case where actions are from a continuous space, we construct a new neural network to represent a deterministic actor function $\mu:S \to A$. With the deterministic actor function, we reduce the inner expectation and modify the Bellman equation~(\ref{eq:bellman}) as follows:

\begin{equation}
	Q^\mu = \mathbb{E}_{r_t, s_{t+1}}[r(s_t, a_t) + \gamma Q^\mu(s_{t+1}, \mu(s_{t+1}))].
\end{equation}

Lillicrap {\em et al.}~\cite{lillicrap2015continuous} used an actor-critic approach to optimize an actor-value function and actor policy, based on the Deterministic Policy Gradient (DPG) method by Silver {\em et al.}~\cite{silver2014deterministic}.

The Q function as a critic network is still optimized by minimizing the loss defined in Equation (\ref{eq:Q_loss}). Let $\theta^\mu$ be the parameters of the actor network. Silver {\em et al.}~\cite{silver2014deterministic} proved that the actor network can be updated by applying the policy gradient as in equation (\ref{eq:policy_gradient_}):

\begin{equation}
\label{eq:policy_gradient_}
	\nabla_{\theta^\mu} J \approx \mathbb{E}_{s_t}[ \nabla_{\theta^\mu}Q(s_t, a|\theta^Q) |_{a=\mu(s_t|\theta^\mu)} ].
\end{equation}
Following the chain rule, we have:
\begin{equation}
\label{eq:policy_gradient__}
	\nabla_{\theta^\mu} J \approx \mathbb{E}_{s_t}[ \nabla_{a}Q(s_t, a|\theta^Q) |_{a=\mu(s_t|\theta^\mu)} \nabla_{\theta^\mu}\mu(s_t | \theta^\mu)].
\end{equation}

In practice, direct implementation of equation \ref{eq:Q_loss} and \ref{eq:policy_gradient__} with neural networks has proven unstable in many cases. Lillicrap {\em et al.}~\cite{lillicrap2015continuous} addressed this problem by adding target networks for both critic and actor as inspired by Minh {\em et al.}~\cite{mnih2013playing}. The parameters $\theta'$ of target networks are updated by slowly tracking the learned networks: $\theta' = \tau \theta + (1-\tau)\theta'$, where $\tau \ll 1$ is a scalar factor.

\begin{algorithm}
\caption{Supervised deep reinforcement learning}
\label{algorithm}
\begin{algorithmic}[1]
\State Randomly initialize critic $Q(s,a|\theta^Q)$, actor $\mu_a(s|\theta^\mu)$ and pioneer $\mu_p(s|\theta^\mu_{t_2})$ networks.
\State Initialize target network $Q'$ and $\mu_a'$ with parameters $\theta^{Q'} \gets \theta^Q$ and $\theta^{\mu'} \gets \theta^\mu$
\State Initialize replay buffer $R$ and $R_p$
\State Initialize pioneer buffer threshold $r_p$
\State Initialize $k \gets 1$
\For{episode = 1 to M}
	\State Initialize a random process $\mathcal{N}$ for exploration.
	\State Initialize temporary buffer $R_e$
	\State Reset environment and receive initial observation state $s_1$.
	\For{t = 1, T}
		\State Select action $a_t = \mu(s | \theta^\mu)$ and execute
		\State Observe reward $r_t$ and new state $s_{t+1}$
		\State Store transition $(s_t, a_t, r_t, s_{t+1})$ in $R$ and $R_e$
		\State Sample N transitions $(s_i, a_i, r_i, s_{i+1})$ from $R$
		\State Set $y_i = r_i + \gamma Q'(s_{i+1}, \mu'(s_{i+1})|\theta^{Q'})$
		\State Update critic by minimizing loss:
		\State ~~~~$L = \frac{1}{N}\sum_i(y_i - Q(s_i, a_i|\theta^Q))$
		\State Update the actor using sampled policy gradient:
		\State $\nabla_{\theta^\mu} J \approx \frac{1}{N}\{(1-k)\sum_i \nabla_{a}Q(s_i,a_i) \nabla_{\theta^\mu}\mu_a(s_i)$ \\~~~~~~~~~~~~~~~~~~~~~~~~$+k\sum_i[\mu(s_i)-a_i]\} $
		\State Update the target networks:
		\State ~~~~~~$\theta^{Q'} = \tau\theta^Q + (1-\tau)\theta^{Q'}$
		\State ~~~~~~$\theta^{\mu'} = \tau\theta^\mu + (1-\tau)\theta^{\mu'}$
	\EndFor
	\If{episode total reward $\ge r_p$}
		\State Move transitions from $R_e$ to $R_p$
	\EndIf
	\State Copy the learning network to pioneer: $\theta^\mu_{t_2} \gets \theta^\mu$
	\State Sample transaction $(s_j, a_j, r_j, s_{j+1})$ from $R_p$
		\State Update the pioneer policy by applying gradient:
		\State ~~~~~~$\nabla_{\theta^\mu_{t_2}} J_p = \sum_j[\mu_p(s_j|\theta^\mu_{t_2}) - a_j]$

	\State Empty $R_e$
	\State Decrease $k$ if a target score is achieved.
	\State Re-initialize the learning network $\theta^\mu \gets \theta^\mu_{t_2}$ if $k$ is decreased.
	\State Increase $r_p$
\EndFor
\end{algorithmic}
\end{algorithm}

\section{Our method}

\begin{figure}[t!]
   \centering
   \includegraphics[width=0.45\textwidth]{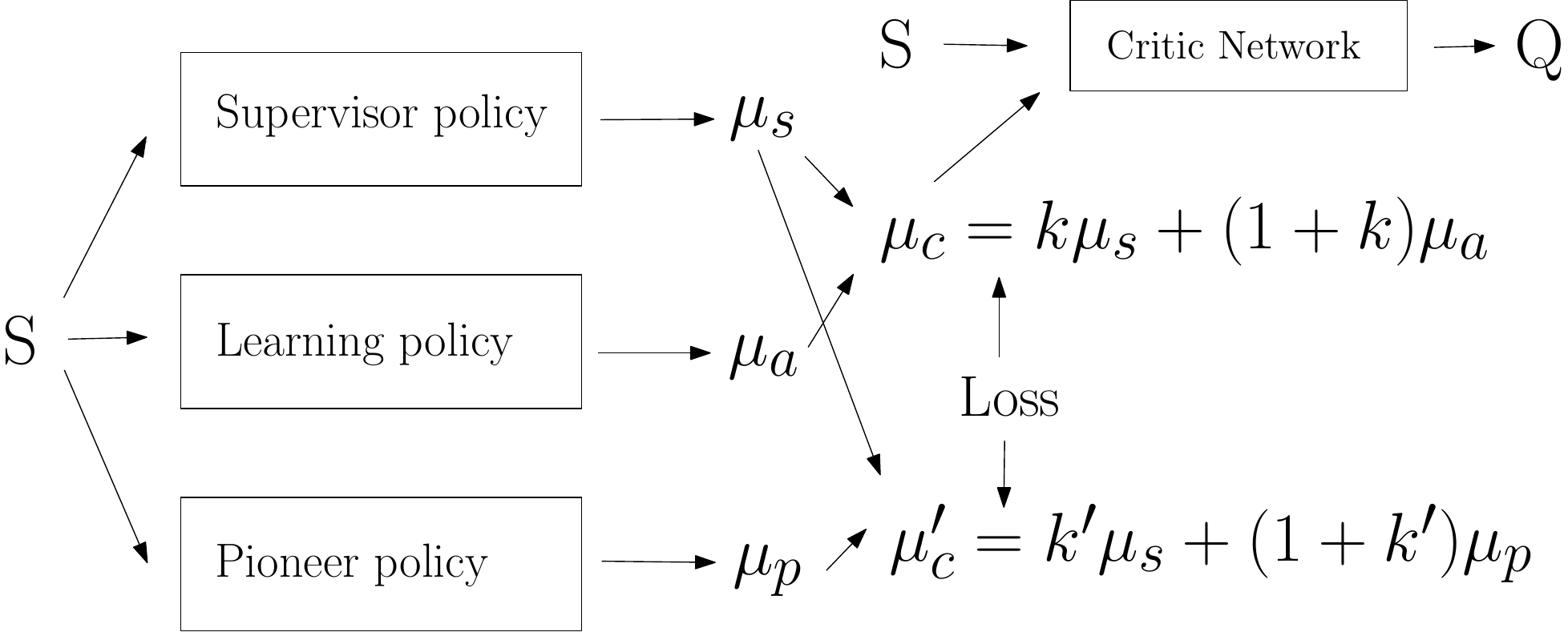}
   \caption{The framework: based on the actor-critic framework, we introduce a supervisor policy, a learning policy, and a pioneer policy. The combined action $\mu_c$ is executed. The pioneer policy is trained such that $\mu_c'$ outputs similar results as $\mu_c$ with an updated weight $k$. The learning policy is then replaced by $\mu_p$.}
   \label{fig:flow}
\end{figure}

Learning from scratch using the algorithms discussed above is impractical in many real world tasks.  The large number of trials and errors can be destructive for agents like unmanned aerial vehicle and many other robots. In a real world learning process, however, supervisors significantly improve the performance of a learner and reduce the number of failures.

Inspired by real world learning experience, we introduce a supervisor into the reinforcement learning process. This section gives detailed explanation of how to combine a supervisor policy and a learning network to train the network and perform tasks at the same time. We slowly reduce and eventually remove the supervisor contribution throughout the learning process.

Let supervisor $\mu_s:S\to A$ be a deterministic actor function, $\mu_a:S\to A$ be our learning policy. We combine the two policies plus some exploration as a new policy, called the {\em combined policy}:
\begin{equation}
\mu(s) = k \mu_s(s) + (1-k) \mu_a(s) + \mathcal{N}_t
\end{equation}
where the {\em combination factor} $k\in [0,1]$ is a scalar and $\mathcal{N}_t$ is a noise process that diminishes as time increases.

In our proposed algorithm, the critic function $Q(s, a| \theta^Q)$ is also modeled by an artificial neural network with parameters $\theta^Q$, the actor network $\mu_a(s|\theta^\mu)$ is modeled by another multi-layer neural network, where $\theta^\mu$  is the parameters of the actor network. Because $\mu_s$ is a fixed policy, $\theta^\mu$ is also the parameters of the combined actor network $\mu(s|\theta^\mu)$.

\subsection{Improving the combined actor}
Assuming $k$ is a constant factor, we consider how to improve the combined policy during learning.

The critic network parameters $\theta^Q$ are optimized by minimizing the loss as defined in equation~(\ref{eq:Q_loss}). The actor network parameters are updated by applying gradient defined in equation~(\ref{eq:policy_gradient_}). But after applying the chain rule, because $\mu$ is a combined policy, equation~(\ref{eq:policy_gradient_}) becomes:
\begin{equation}
\label{eq:rl_policy_gradient}
\begin{split}
\nabla_{\theta^\mu} J_a & \approx \mathbb{E}_{s_t}[ \nabla_{\theta^\mu}Q(s_t, a|\theta^Q) |_{a=\mu(s_t|\theta^\mu)} ] \\
& = \mathbb{E}_{s_t}[ \nabla_{a}Q(s_t, a|\theta^Q) |_{a=\mu(s_t|\theta^\mu)} \nabla_{\theta^\mu}\mu(s_t | \theta^\mu)] \\
& = \mathbb{E}_{s_t}[ \nabla_{a}Q(s_t, a|\theta^Q) |_{a=\mu(s_t|\theta^\mu)} \cdot \nabla_{\theta^\mu}\mu_a(s_t | \theta^\mu)] \\
&~~~\cdot (1-k)
\end{split}
\end{equation}

Equation~(\ref{eq:rl_policy_gradient}) considers only the performance improvement of the combined policy. To better train the learning policy, we also want the learning policy to behave as closely as possible to the supervisor. The parameters $\theta^\mu$ of the learning network can also be updated by minimizing the loss:
\begin{equation}
L(\theta^\mu) =  \frac{1}{2} \mathbb{E}[(\mu_a(s|\theta^\mu) - \mu_s(s))^2],
\end{equation}
whose corresponding gradient is:
\begin{equation}
\nabla_{\theta^\mu} J_s = \mathbb{E}[ \mu_a(s|\theta^\mu) - \mu_s(s) ].
\end{equation}

In order to improve the performance of a combined network and learn from the supervisor at the same time, we apply a combined gradient to the learning actor network, as defined in the following equation
\begin{equation}
\label{eq:policy_grad}
\nabla_{\theta^\mu} J = \nabla_{\theta^\mu} J_a + \lambda \nabla_{\theta^\mu} J_s,
\end{equation}
where $\lambda$ is a scalar factor.

\subsection{Reducing supervision}
We now consider reducing the contribution of the supervisor to our combined policy.

Let $k_{t_1}$ be the value of $k$ at time $t_1$. We choose $k_{t_2}$ to be no larger than $k_{t_1}$ for any $t_2 \ge t_1$. Notice that when $k \to 0$, the combined actor is the learning actor. However, equation~(\ref{eq:policy_grad}) prevents the learning actor from being much better than the supervisor if $\lambda$ is a constant number. Assuming the learning actor has learned the supervisor policy, we set $\lambda \gets k$ so the learning network can improve without relying on the supervisor.

In practice, we update the $k$ value after achiving a target score for one or more epochs, so the combined policy has more time to be improved and stablize. Let $\mu_{t_1}$ be the combined policy at time $t_1$, $k_{t_1}$ be the $k$ value at time $t_1$. Assuming that at time $t_2 > t_1$, $\mu_{t_1}$ is well trained and $\mu_{t_2} \neq \mu_{t_1}$, shifting the combined policy from $\mu_{t_1}$ to $\mu_{t_2}$ can result in bad performance.

We address the problem by adding a copy of the learning network $\mu_p(s | \theta^\mu_{t_2})$, named the {\em pioneer} network, representing the learning actor at time $t_2$, where $\theta^\mu_{t_2}$ are the parameters. Knowing the value of $k_{t_2}$ and parameters $\theta^\mu_{t_1}$ of the learning actor at time $t_1$, the combined policy at the time-step $t_2$ is known:
\begin{equation}
\mu_p(s_{t_2} | \theta^\mu_{t_2}) = k_{t_2} \mu_s(s_{t_2}) + (1-k_{t_2}) \mu_a(s_{t_2}| \theta^\mu_{t_2}) + \mathcal{N}_{t+1}.
\end{equation}

Before shifting the combined policy from $\mu_{t_1}$ to $\mu_{t_2}$, we optimize the pioneer network such that it behaves as similarly to previous combined policies as possible. Parameters $\theta^\mu_{t_2}$ are updated by minimizing
\begin{equation}
L(\theta^\mu_{t_2}) = \frac{1}{2} \mathbb{E}[ (\mu_p(s) - a)^2 ]
\end{equation}
where $a$ is the output by previous combined policies. This is equivalent to applying gradients
\begin{equation}
\label{eq:pioneer_grad}
\nabla_{\theta^\mu_{t_2}} J_p = \mathbb{E}[\mu_p(s) - a]
\end{equation}
to the pioneer network.

Applying gradients (\ref{eq:pioneer_grad}) using randomly sampled previous transitions requires previous combined policies to have stably good performance. In practice, this requirement is not always satisfied. We use a priority replay buffer to store and sample state-action pairs with high returns.

In our implementation, we adapt the idea of using target networks from Lillicrap {\em et al.}~\cite{lillicrap2015continuous} and Minh {\em et al.}~\cite{mnih2013playing}, to prevent actor and critic networks from divergence. Algorithm~\ref{algorithm} is the pseudo-code of our algorithm.

When the combination factor $k$ gets to zero, our method is reduced to DDPG algorithm. So the learning policy is guaranteed to improve its performance over more trails.

\section{Experiments}
In this section, we test our method under several OpenAI Gym~\cite{brockman2016openai} environments and discuss the impact of pioneer network and supervisors with different performances. Our primary environment is LunarLanderContinuous-v2, which operates a landing agent by setting main engine and side engine forces to land on a pad centered at $(0,0)$. In this environment, if an episode's total return is less than 0, the agent has crashed, if the return is larger than 200 the agent is landed successfully, otherwise, the agent has landed but not in the desired range.

\subsection{Pioneer network}
\label{sec:pioneer}
We also performed an experiment to test the difference of adding a pioneer network. We first trained a supervisor using DDPG method. After 1500 episodes of training, the supervisor performance is relatively stable. Its 100-episode average return is above 200. Our learning network has the exact same structure as the supervisor. We reduced the combination factor $k$ by $6\%$ for every 4 episodes. After 300 episodes, the supervisor contribution was less than $1\%$.

\begin{figure}[h]
   \centering
   \includegraphics[width=0.475\textwidth]{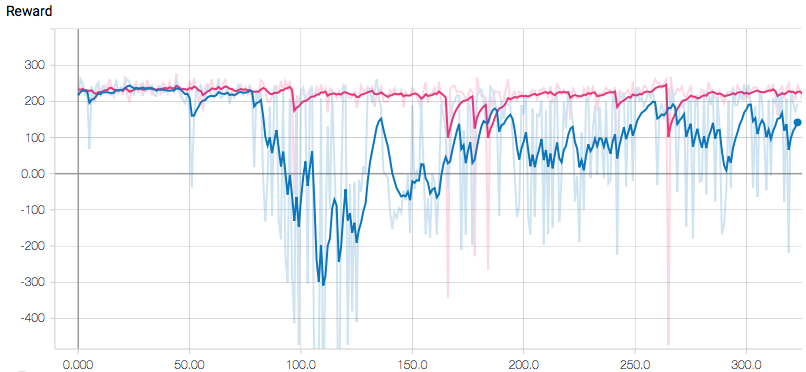}
   \caption{Comparison of the combined actor performances with and without the pioneer network. The x-axis is the number of episodes and the y-axis is the total reward in one episode. The pink curve shows the performance with the pioneer network while the blue curve is without the pioneer. Data smoothed for visual purposes. }
   \label{fig:pioneer}
\end{figure}

Figure~\ref{fig:pioneer} shows the comparison of the combined actor performance with and without the pioneer network. Without a pioneer network, the combined actor performance drops sharply when the combination factor $k$ is less than 0.47. Then performance improves throughout the remainder of training. With the pioneer network, the performance is much more stable (average reward larger than 200) and the learning process is much faster than when learning from scratch. This indicates the pioneer network significantly improves the stability of the learning process.

Although there are still crashes (reward below 0) with the pioneer added, we suspect this is caused by the imperfect supervisor; running the supervisor alone also leads to some crashes.

\subsection{Supervisor impact}
The next experiment is for testing the combined actor performance under different supervisors. Our supervisors are trained as in Section~\ref{sec:pioneer}. We pick two supervisors: one (bad supervisor) trained for 1000 episodes whose 100-episode average performance is around 100  and one (good supervisor) trained for 1500 episodes whose 100-episode average performance is above 200. We enable the pioneer network in both experiments and reduce the combination factor $k$ by $6\%$ for every 4 episodes.

\begin{figure}[h]
   \centering
   \includegraphics[width=0.48\textwidth]{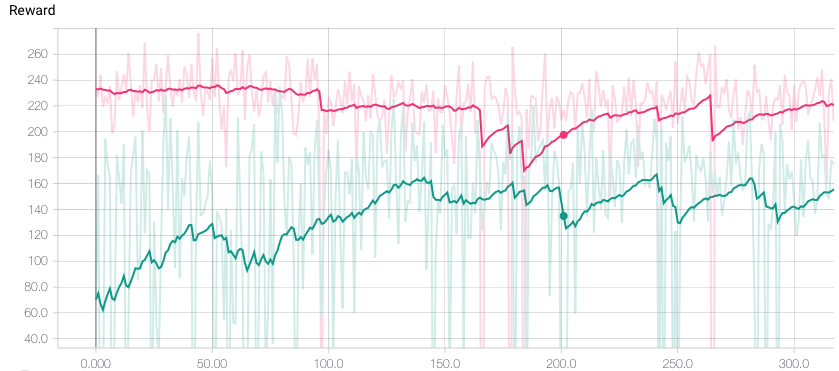}
   \caption{Comparison of combined actor performance trained under supervision of both a good (pink) and bad (green) policies. Data smoothed for visualization.}
   \label{fig:bad_vs_good}
\end{figure}

Figure~\ref{fig:bad_vs_good} shows the comparison of learning from good and bad supervisors. When learning from the good supervisor, the combined actor performance is stably good (rewards mostly above 200). However, with a bad supervisor, after 300 episodes, the rewards are between 130 and 160 and improve slowly.

\begin{figure}[h]
   \centering
   \includegraphics[width=0.48\textwidth]{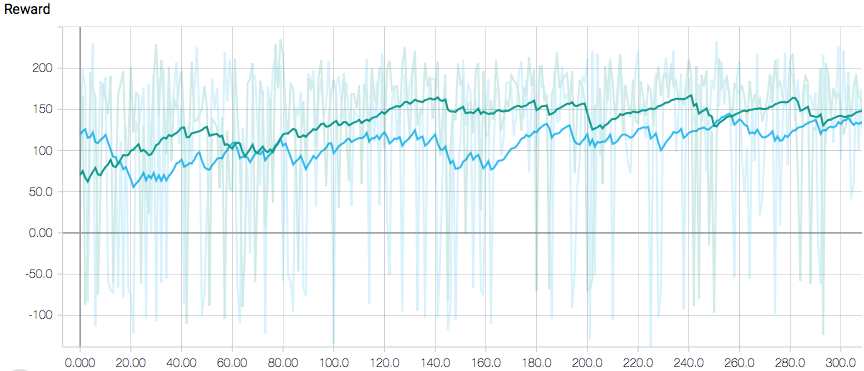}
   \caption{Network performance under the supervision of a bad supervisor. The blue curve is the performance of the supervisor alone and the green curve is the performance of the combined actor. }
   \label{fig:bad}
\end{figure}

Figure~\ref{fig:bad} shows the comparison of our combined actor and the supervisor. The combined actor performs better than the supervisor alone. In 300 episodes, the supervisor crashed 54 times and successfully landed 30 times, while the combined actor crashed only 23 times and successfully landed 37 times.

Figure~\ref{fig:bad} indicates our method is learning and improving the performance of a given supervisor.

\begin{figure}[h]
   \centering
   \includegraphics[width=0.48\textwidth]{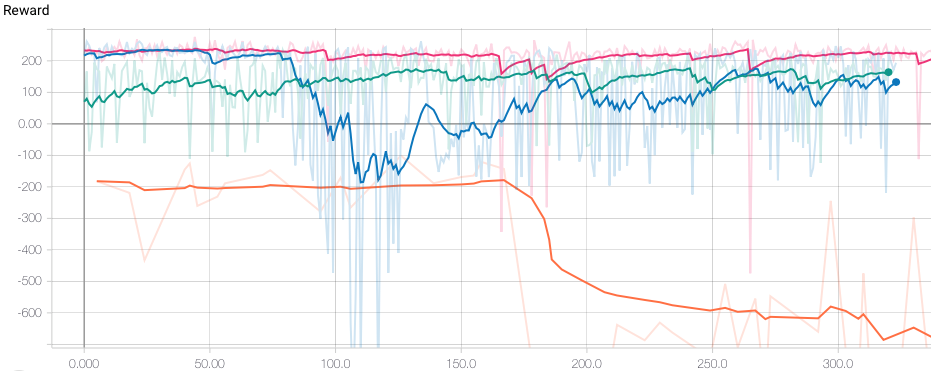}
   \caption{Performance curves in the first 340 episodes for the DDPG method (orange), our proposed method with a good supervisor (pink), our method without pioneer network (blue) and our method using a bad performance supervisor (green). }
   \label{fig:lunar_all}
\end{figure}

Figure~\ref{fig:lunar_all} shows comparison of performances in the first 340 episodes for different methods. The DDPG algorithm achieved no successful landing, our method with a good supervisor is consistently landing with success, and without pioneer network, the performance over the training process is very unstable.

\subsection{Mujoco environments}
We also ran several Mujoco environments to test our algorithm in different continuous control tasks under supervisors with different performance qualities.

For the InvertedPendulum-v1(Figure~\ref{fig:pendulum}) environment the goal is to swing up a pendulum and make it stand as long as possible. Getting a reward of higher than 950 is considered ``solved''. We also use DDPG algorithm to train two supervisors: a bad one capable of getting rewards around 255 and a good one capable of getting rewards over 950. We decrease the combination factor by $10\%$ for every 5 episodes until supervisor is contributing less than $1\%$.

\begin{figure}[h]
   \centering
   \includegraphics[width=0.475\textwidth]{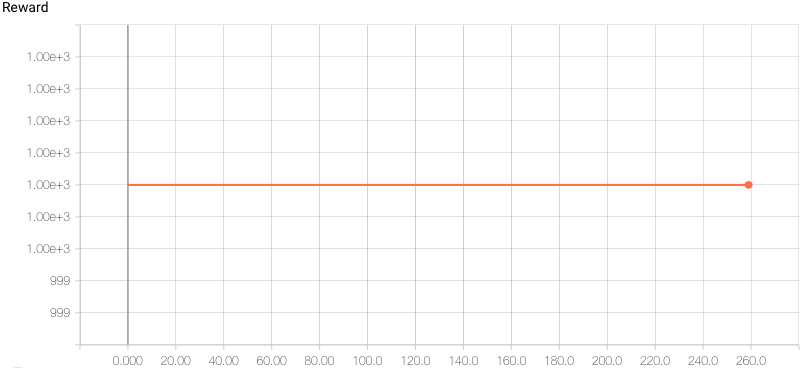}
   \caption{Training to stand a pendulum with a good supervisor. Our method is achieving rewards of 1000 over the whole process.}
   \label{fig:pend_good}
\end{figure}

The result of training with a good supervisor is shown in Figure~\ref{fig:pend_good}. With a good supervisor, our combined policy gets the highest rewards (1000) over all trails.

However, training with a bad supervisor is only slightly better than using the supervisor alone in this environment. We believe the reason is that, in this specific task, the distribution of states under a good supervisor policy is very concentrated in a small region where the pendulum is close to stand. A bad supervisor generates states distributed much wider in the state space; it seems a lot of effort is spent on states that are less important. This experiment indicates that if a supervisor policy is not good enough, the learning policy will not improve significantly. The next experiment on HalfCheetah-v1 environment uses a supervisor that is close to solve the task.

\begin{figure}[h]
   \centering
   \includegraphics[width=0.475\textwidth]{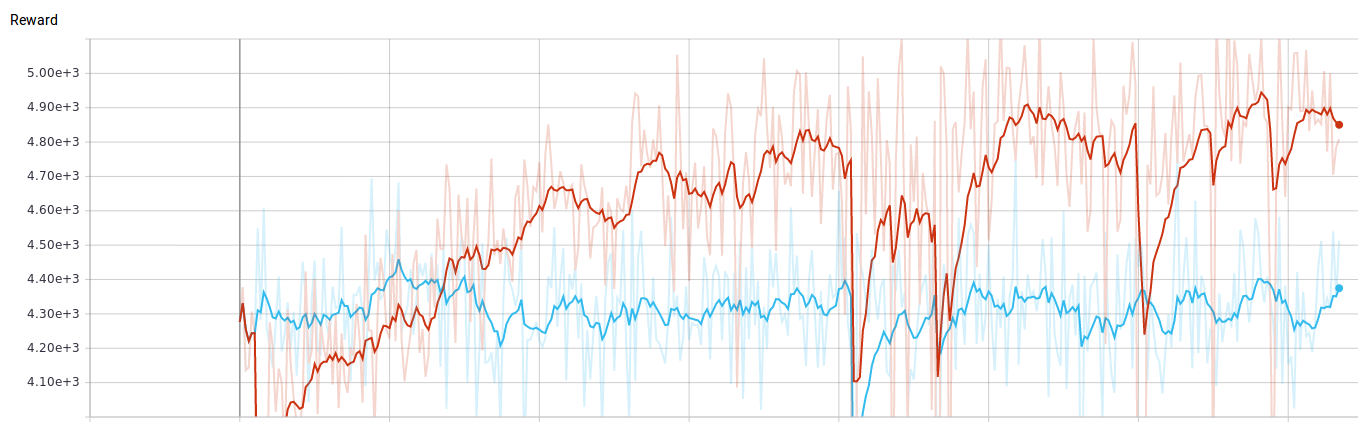}
   \caption{Training to run a 2D cheetah robot with a bad supervisor. Our combined actor is graduately performing better than the supervisor and finally achieved a target score of 4800.}
   \label{fig:half_cheetah_bad}
\end{figure}

The goal for the HalfCheetah-v1(Figure~\ref{fig:half_cheetah}) environment is to control actions of a 2D cheetah robot and keep it running as long as possible. Getting a reward of higher than 4800 is considered ``solved''. We trained using a bad supervisor that gains rewards around 4300. This supervisor was trained for over 5000 epoches, and not able to improve for 10000 more epoches, indicating a neural network not complicated enough. Performance is shown in Figure~\ref{fig:half_cheetah_bad}

\begin{figure}[h]
   \centering
   \includegraphics[width=0.475\textwidth]{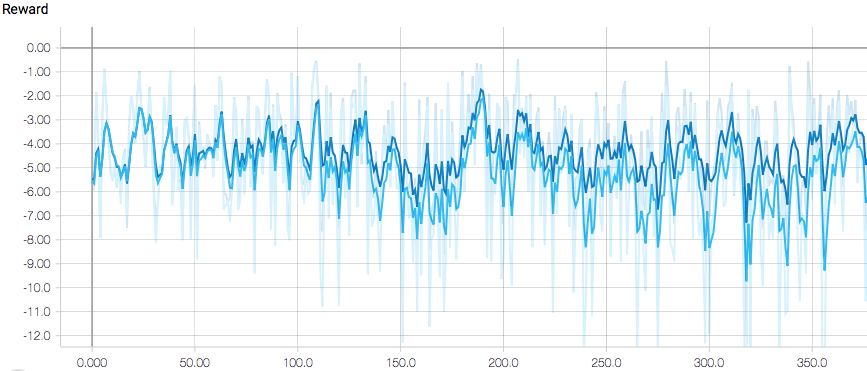}
   \caption{Reacher-v1 environment. Our combined actor (light blue) is performing similar as the supervisor (dark blue).}
   \label{fig:reacher_good}
\end{figure}

We also run our method on Reacher-v1 (Figure~\ref{fig:reacher}) environment. The goal for this environment is to control a 2R arm to reach and stay in a goal position. Getting a reward of higher than -3.75 is considered ``solved''. We used a supervisor policy that gains on average rewards of -5, which is still far from solving the task in this environment. Our combined actor is getting similar performance. See Figure~\ref{fig:reacher_good}. Considering that training this supervisor took over 10000 episodes. Having a similar near-solution policy trained in 350 episodes is not bad. Our explination for this experiment is the same as in the InvertedPendulum-v1 experiment, if a supervisor is too far from acheiving a good score, the learning policy will also have hard time to make good improvement.

\section{Conclusion}
In this work, we introduced an algorithm to safely and quickly learn from a supervisor policy and eventually improve performance. We combined a supervisor with a learning network, and gradually decay the contribution of supervisor to force the learning network to learn from previous executions. By introducing a pioneer network, we are able to stabilize the combined policy performance over the whole learning process.

The proposed method can be considered as an efficient way of reducing exploration and exploitation on state-action pair space. Simulated experiments proved the efficiency of the algorithm; next steps include applying this algorithm to real-world robot systems.

\addtolength{\textheight}{-12cm}   


\bibliographystyle{plain}
\bibliography{references}

\end{document}